\documentclass[runningheads]{llncs}
\usepackage{wasysym}
\usepackage{booktabs}
\usepackage{graphicx}
\usepackage{color}
\usepackage{siunitx}
\usepackage[breaklinks=true]{hyperref}
\hypersetup{colorlinks=true}
\usepackage{xurl}
\usepackage{subcaption}
\newcommand{\mjb}{MJbots}

\begin{document}
\title{TurtleRabbit 2024 SSL Team Description Paper}
%
%
\author{Linh Trinh\thanks{Authorship: team lead first, contributors in alphabetical order, academic lead last.} \and Alif Anzuman \and Eric Batkhuu \and Dychen Chan \and Lisa Graf\thanks{Lisa Graf is with the Neurorobotics Lab, Universit\"at Freiburg, Germany, and was a visiting researcher at Western Sydney University, Jan/Feb 2024.} \and Darpan Gurung \and Tharunimm Jamal \and Jigme Namgyal \and Jason Ng \and Wing Lam Tsang \and X.~Rosalind Wang\orcidID{0000-0001-5454-6197} \and Eren Yilmaz \and Oliver Obst\orcidID{0000-0002-8284-2062}}
\authorrunning{L. Trinh et al.}
%
\institute{Western Sydney University, Locked Bag 1797. Penrith NSW 2751, Australia\\
Website \url{https://wsu-turtlerabbit.github.io/}\\
Corresponding author: \email{o.obst@westernsydney.edu.au}}
\maketitle              
%




\begin{abstract}
TurtleRabbit is a new RoboCup SSL team from Western Sydney University. This team description paper presents our approach in navigating some of the challenges in developing a new SSL team from scratch. SSL is dominated by teams with extensive experience and customised equipment that has been developed over many years. Here, we outline our approach in overcoming some of the complexities associated with replicating advanced open-sourced designs and managing the high costs of custom components. Opting for simplicity and cost-effectiveness, our strategy primarily employs off-the-shelf electronics components and ``hobby'' brushless direct current (BLDC) motors, complemented by 3D printing and CNC milling. This approach helped us to streamline the development process and, with our open-sourced hardware design, hopefully will also lower the bar for other teams to enter RoboCup SSL in the future. The paper details the specific hardware choices, their approximate costs, the integration of electronics and mechanics, and the initial steps taken in software development, for our entry into SSL that aims to be simple yet competitive. 


\end{abstract}
\section{RoboCup Small Size League without the custom PCBs}

Our team, TurtleRabbit, aims to qualify as a new participant for the RoboCup 2024 Small Size League (SSL). SSL is one of the oldest leagues in RoboCup, and despite strong participation in its earliest days~\cite{WBT99,CBW01,Howa01}, it has been 20 years~\cite{BWN04} since an Australian team qualified for participation in this league.  
Many of the currently participating institutions have a long history and tremendous experience, which is evident from their extended team descriptions and some of the open-source designs. This experience is reflected in the level of detail with which parts of the designs are created, whether to save space, ease maintenance and repairs, or to ensure longevity and protect the components. While this sophistication in the design of the robots appears to be a factor in their success, it poses significant technical and financial barriers for new teams such as ours in replicating these successful approaches. Creating a team comparable to most of the existing ones requires expertise in fabricating custom circuit boards, mechanics, and software, ranging from low-level controllers to methods that determine high-level team actions.

Contrary to this trend, our strategy is deliberately simple and relies on (mostly) off-the-shelf components. As a result, our robot locomotion hardware is limited to a single Raspberry Pi 4 (RPI), paired with a \mjb{} PiHat and moteus controllers for relatively inexpensive BLDC drone motors. This approach accelerates our development and reduces cost and complexity. For controlling our kicker and dribbler, we employ an Arduino Every board connected to the RPI.

\section{Electronics and motors}
\label{sec:electronics}

Most teams in RoboCup SSL use high-quality BLDC motors for their robots, from electronic manufacturers such as Nanotec, Maxon, Moons, both for driving and dribbling. The relatively high cost of these motors and their controllers can be an obstacle for a new team. We have opted for an approach with cheaper BLDC drone motors, an approach also taken by the GreenTea SSL team~\cite{SOI+23}. For the drive motors, we use a Tarot Martin 4008 brushless drone motor. We use \mjb{} moteus r4 controllers~\cite{Piep24} to turn our drone motors into servo actuators. The moteus boards integrate the necessary drive electronics for 3-phase brushless field-oriented control of BLDC motors, with an STM32G4 microcontroller, an absolute magnetic encoder, and a \SI{5}{Mbps} CAN-FD interface. Moteus boards are commercial, off-the-shelf controllers with an open source firmware, originally developed for quadruped robots. Their on-board absolute encoder has been placed close to a diametrically magnetised magnet that we mount to the secondary output (rear motor axis) with a small 3D-printed holder. 
Moteus boards support position, torque, and velocity control modes. For our robots, we use the boards in velocity control mode. 

The controller boards support communication with a CAN interface with flexible data-rate (CAN-FD), and are connected to a Raspberry Pi 4b with an \mjb{} pi3hat. The pi3hat attaches to the Raspberry Pi GPIO and provides 5 independent CAN-FD interfaces. The pi3hat also powers the Raspberry Pi and contains a \SI{1}{khz} Attitude Reference System / Inertial Measurement Unit (IMU); in its current stage our software does not make use of the IMU yet though we plan to integrate the data into our world model for the 2024 competition. Similar to the moteus boards, the \mjb{} pi3hat firmware is also open source.

To distribute power from a 6S1P Lithium Polymer battery (Lipo) to motors and controllers we use an \mjb{} power distribution board that also pre-charges high capacitance loads like BLDC motors to protect the controllers, and allows software-controlled shut down. 

For the qualification stage, we integrated a single (straight) kicker device using an off-the-shelf push-pull solenoid powered using a DC-DC boost converter to increase its strength, and controlled using an Arduino Nano Every. The Arduino is connected to the Raspberry Pi over USB. The solenoid provides a reasonably strong ``kick'', though custom designed kickers (see, e.g., \cite{ABBB23,BEG+23}) appear to be advantageous. Our design also leaves sufficient space for a secondary chip-kicker but is currently not integrated into our robots. 

Our dribble device is currently still under development and being tested. In order to save costs, we opted for a BLDC micro drone motor, controlled from the Arduino Every using a \SI{20}{A} XRotor ESC. The dribbler is powered by a separate 2S1P Lipo.

An overview of the electronic components along with their approximate prices converted to US\$ can be found in Table~\ref{tab:electronics}. Additionally required are connectors and cables for power, USB and CAN bus, and potentially additional onboard sensors, such as a camera or other sensors for ball detection, if not included with the Raspberry Pi.

The components we chose are mostly off-the-shelf and allowed us to develop the robot without having to create custom PCBs and without designing custom control boards. One potential disadvantage of cheaper drone motors compared to the more ``professional'' BLDC motors used by many other teams is their higher cogging torque, resulting in less smooth motion at low RPM. Apart from simply using faster speeds, a solution to this problem is to adjust moteus controller board parameters in order to compensate for cogging torque (resulting in an overall lower torque bandwidth). So far we have not found this adjustment necessary. 

\begin{table}[htp]
\caption{Electronic components used per robot. Prices are approximate and per item. Total electronics costs for a single robot adds up to approx. US\$1,045 (or AU\$1,600).}
\centering
\begin{tabular}{lllr}
\toprule
\multicolumn{2}{l}{Component} & Specs & Price (US\$) \\
\midrule
($4\times$) & Tarot 4008 Martin BLDC Motor ~ & \SI{85}{g} weight, \SI{44.5}{mm} diameter,  & 55.00\\
& (18N/24 Pole \SI{330}{KV}) &  \SI{22}{mm} height, \SI{497}{W} power  & \\
& & \SI{30}{A} max continuous current & \\
($4\times$) & \mjb{} moteus controller & 3 phase brushless FOC based & 75.00\\
& &  control, 10-\SI{44}{V}, \SI{500}{W} peak& \\ 
& &  power, \SI{100}{A} peak phase current & \\
\multicolumn{2}{l}{\mjb{} pi3hat r4} & $4\times  \SI{5}{Mbps}$ CAN-FD bus, $1\times$ &150.00  \\
& & \SI{125}{kbps} CAN, \SI{1}{kHz} IMU, I2C & \\
\multicolumn{2}{l}{\mjb{} power dist board r4} & CAN-FD bus, energy monitoring &150.00  \\
\multicolumn{2}{l}{Raspberry Pi 4 Model B} & \SI{4}{GB}, \SI{32}{GB} uSD card &80.00  \\
\multicolumn{2}{l}{Arduino Nano Every} &  &16.50  \\
\multicolumn{2}{l}{RCinpower GTS V3 1003 micro motor} & \SI{10000}{KV}, \SI{3.45}{g} weight& 15.00 \\
\multicolumn{2}{l}{XRotor ESC} & \SI{20}{A}, 2S-4S& 15.00 \\
($2\times$)&  DC-DC Boost Converter & 45-\SI{390}{V} output &5.00  \\
& (we currently use only one) & & \\
($2\times$)& Push-pull solenoid (Adafruit) & \SI{144}{g} weight & 25.00  \\
& (we currently use only one) & & \\
\multicolumn{2}{l}{6S1P Lipo Battery CNHL Ministar} & \SI{1000}{mAh} &45.00  \\
\multicolumn{2}{l}{2S1P Lipo Battery CNHL Ministar} & \SI{450}{mAh} &24.00  \\
\bottomrule
\end{tabular}
\label{tab:electronics}
\end{table}%

\section{Mechanical design}

\begin{figure}[tbp]
\centering
\includegraphics[height=3cm]{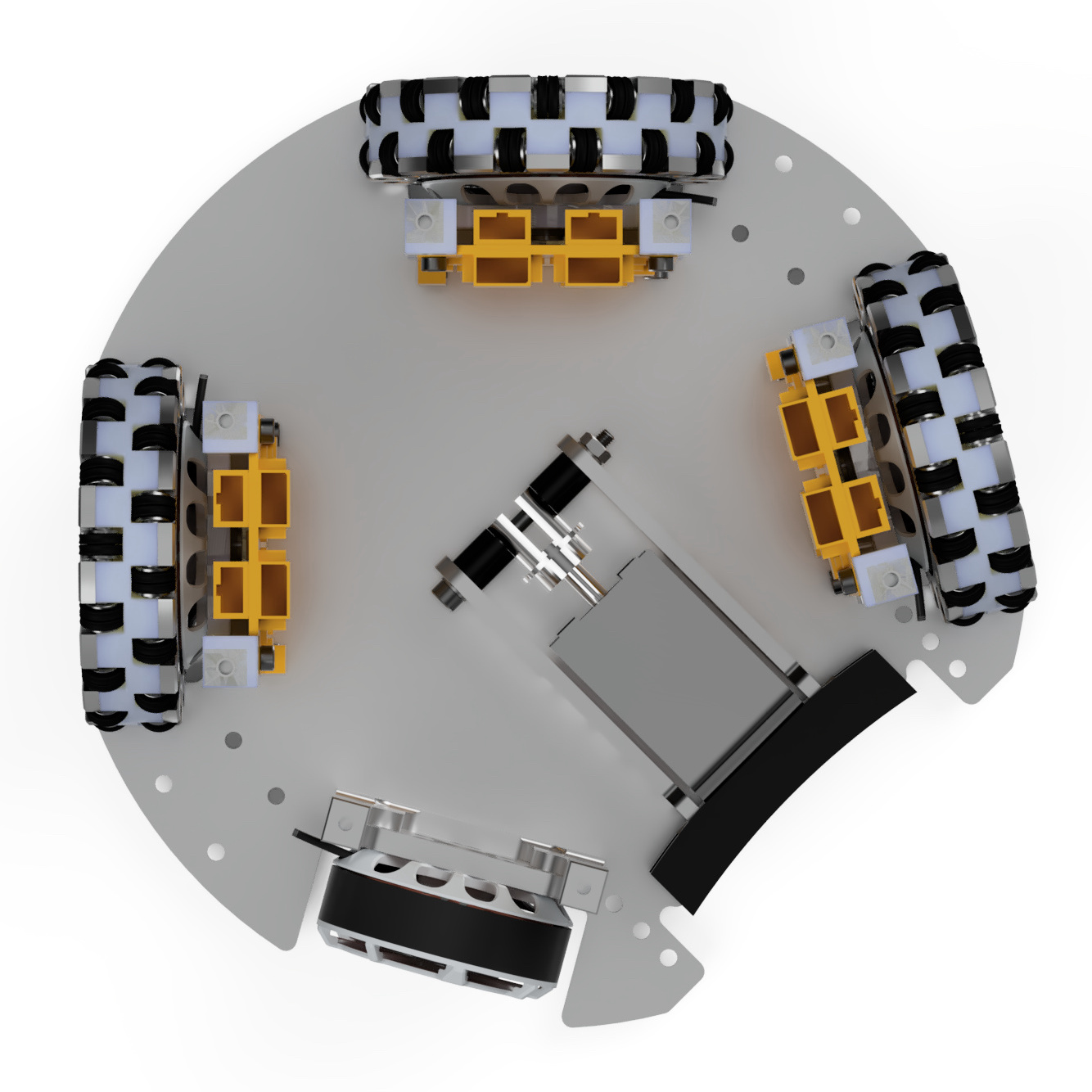}\hfill
\includegraphics[height=3cm]{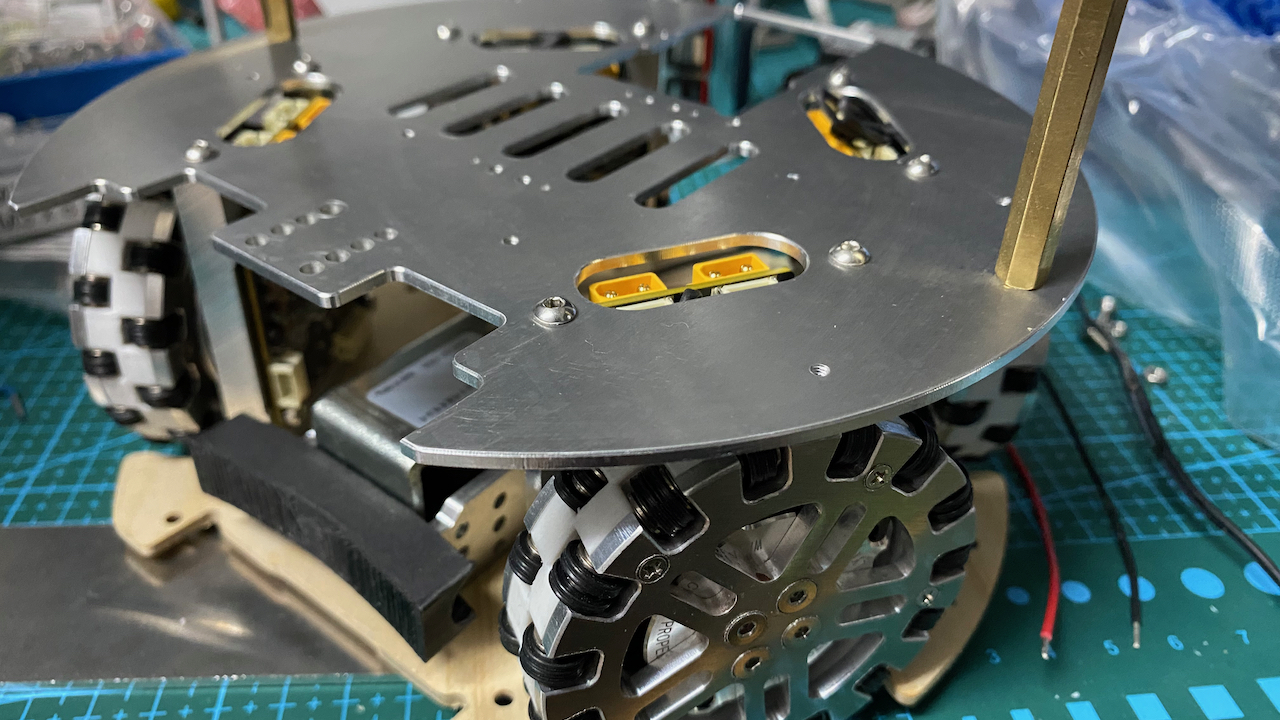}\hfill
\includegraphics[height=3cm]{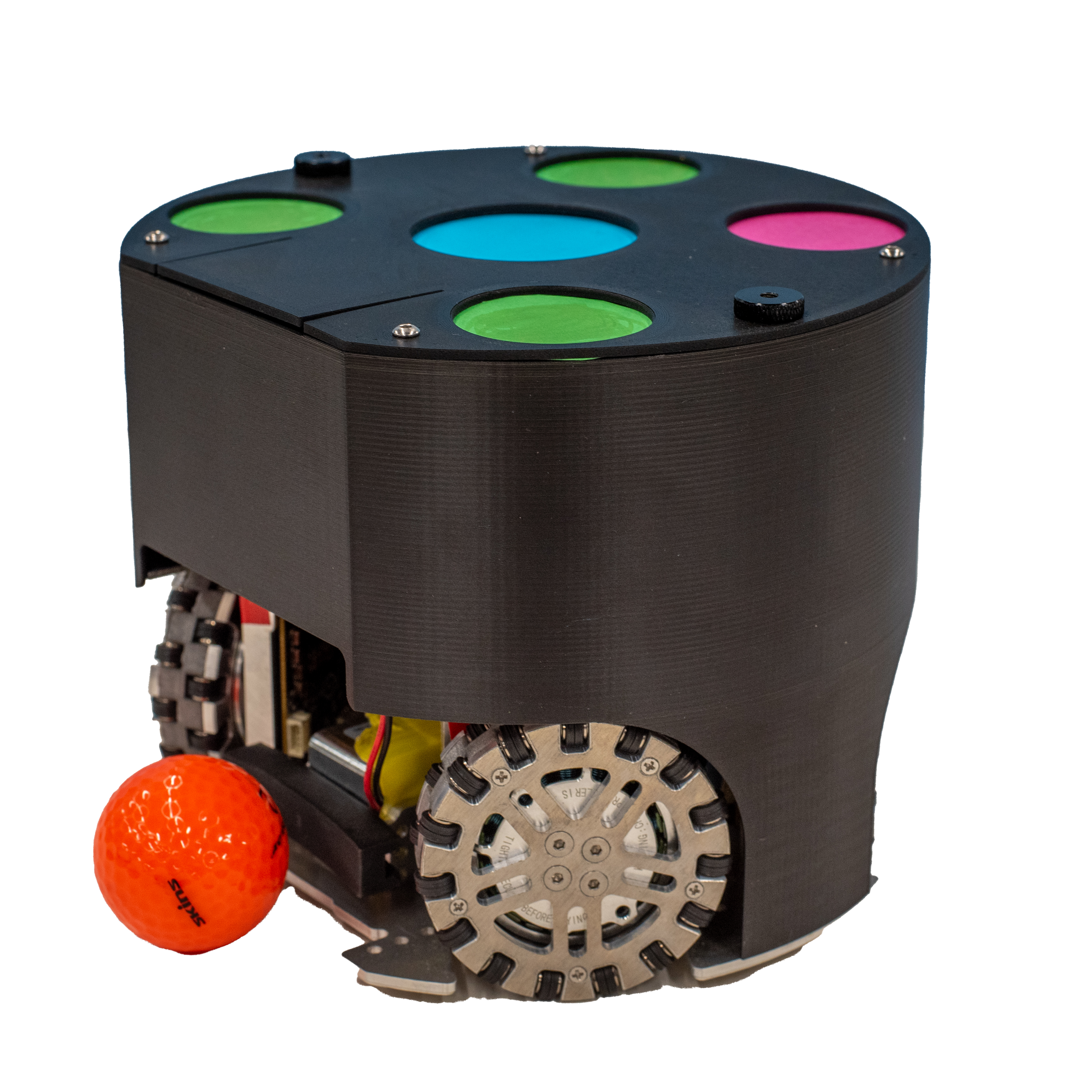}
\caption{Left: Render of the base plate with wheel arrangement and kicker (front right wheel and controller removed, exposing motor). Centre: Intermediate stage, plywood base plate. Right: completed, with PETG shell. }
\label{fig:topview}
\end{figure}

For the initial mechanical design, we investigated both a 3-wheeled and a 4-wheeled setup. While a 3-wheeled setup appeared attractive because of overall lower costs, the more conventional 4-wheel configuration makes it easier to evenly distribute weight and also leaves more space along the longitudinal axis, e.g., for kicker or battery. We kept the two front wheels at a $120^\circ$ angle (as they would be in a 3-wheel configuration) to make space for a kicker and dribbler, and the two rear wheels at a $90^\circ$ angle from each other (see Fig.~\ref{fig:topview} left). Since our motor controllers need to be close to the (relatively flat) drone motors, they have to be mounted vertically resulting in a mechanical design different to most teams with exception of GreenTea~\cite{SOI+23}, though the subsequent sections will also highlight some differences to their approach. Our custom mechanical parts are machined on a Makera Carvera Desktop CNC, or printed on Ender 5+ and Bambu Lab X1-Carbon 3D printers. We used (mostly) Fusion 360 for design; all designs and STEP files are available on our GitHub \href{https://github.com/WSU-TurtleRabbit}{https://github.com/WSU-TurtleRabbit}. 

\subsection{TurtleRabbit Omniwheel}

\begin{figure}[tbp]
\centering
\includegraphics[width=\textwidth]{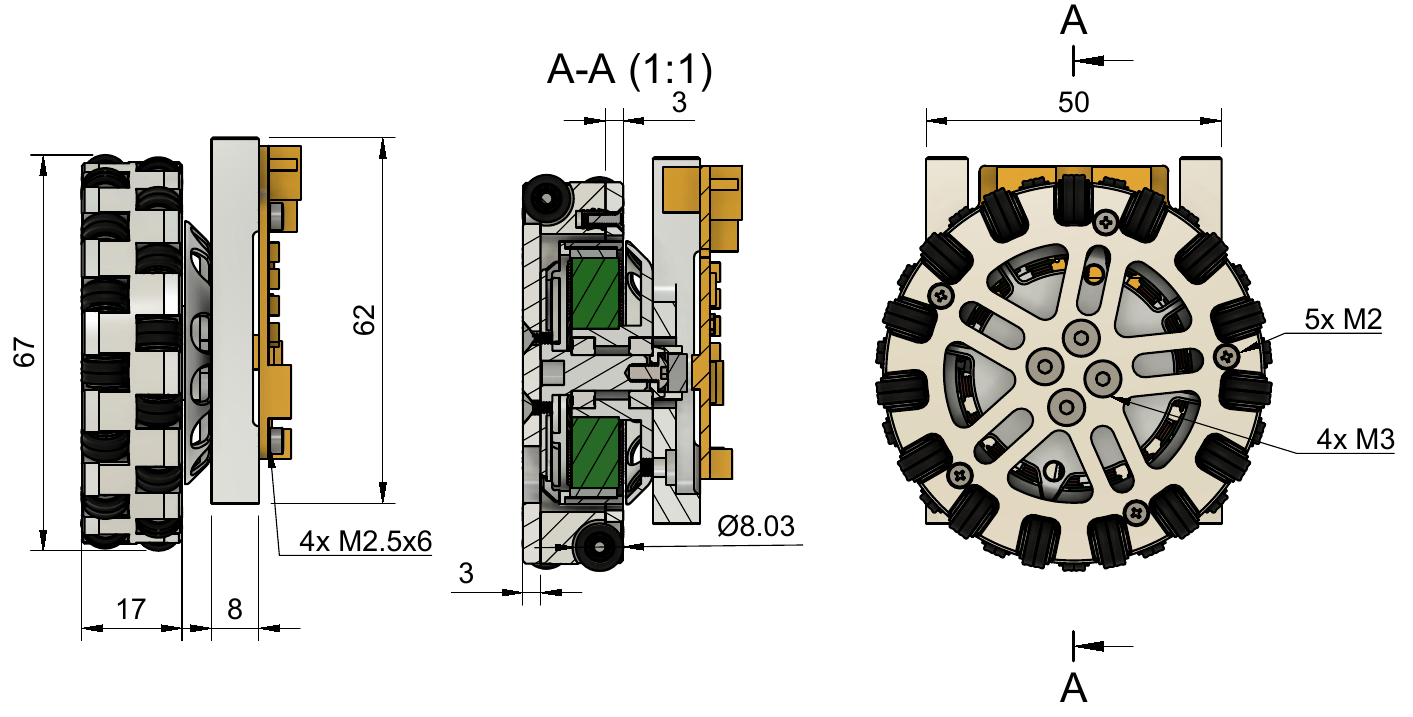}
\caption{The TurtleRabbit omniwheel drive with motor, controller, and motor mount.}
\label{fig:omniwheel}
\end{figure}

The Tarot 4008 Martin motor is an efficient BLDC motor designed for quadcopters. It is a small, ``outrunner'' type motor with a diameter of \SI{44.5}{mm}, and a height of \SI{22}{mm}. In order to save space, we designed the wheels so the motor sits inside the main wheel chassis. As a result, the main wheel chassis is large, comparing to other teams, with a diameter of \SI{64.5}{mm} (\SI{67}{mm} wheel diameter incl.\ subwheels, see Fig.~\ref{fig:omniwheel}).

We based the principal wheel design on GreenTea~\cite{SOI+23}, who use a fully (PLA) printed wheel chassis with a similar sized BLDC motor to ours, but we were also concerned about mechanical and thermal issues (cf.~\cite{RJ20}). As a compromise, we decided to print the main wheel chassis in PETG, while machining both the front and rear wheel covers from \SI{3}{mm} 6061-T6 aluminium. The aluminium front wheel cover, directly attached to the flat front of the BLDC motor, offers thermal conductivity and heat dissipation properties critical in managing the elevated temperatures generated by the motor during operation. We used M2 brass heat-set inserts to provide robust threading for the connection between wheel covers and the wheel chassis.

\subsubsection{Subwheel design}

\begin{figure}[tbp]
\centering
\includegraphics[height=2.92cm]{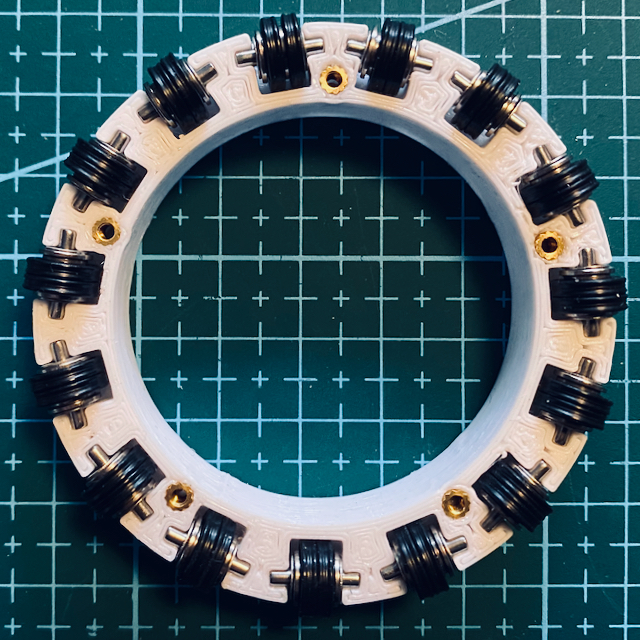}\hfill%
\includegraphics[height=2.92cm]{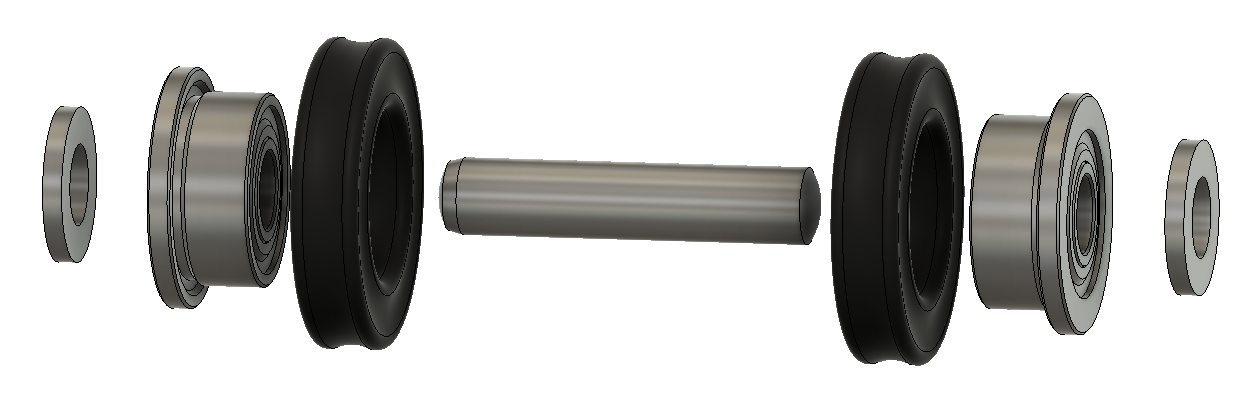}
\caption{Left: Subwheels mounted on PETG wheel chassis. Right: Subwheel exploded view, with washers, MF52 2Z bearings, NBR $\SI{4.47}{mm} \times \SI{1.78}{mm}$ X-ring, $\SI{2}{mm} \times \SI{10}{mm}$ dowel pin.}
\label{fig:subwheel}
\end{figure}

Our subwheel design (Fig.~\ref{fig:subwheel}) is inspired by Tigers Mannheim wheel~\cite{RJ20}, with some differences mostly due to availability or cost of components in Australia. Each main wheel holds 2 rows of 15 subwheels. Each of the 30 subwheels consists of a $\diameter$ \SI{2}{mm} $ \times$ \,\SI{10}{mm} steel dowel pin, two $2\times5\times2.5$ flanged bearings, 2 M$2 \times \SI{0.5}{mm}$ washers, and two NBR X-rings (ID $\SI{4.47}{mm} \times $\SI{1.78}{mm} thickness) as ``tyres''.

\subsection{Motor mounts, base plates, shell and other parts}

Our motor mounts hold the BLDC motor on one side, and the controller on the opposite side. In our design, they also act as the connection between base plate and mid level plate in our design. Initially, we 3D printed the motor mounts, and cut the base plates from \SI{3}{mm} plywood, but for thermal and mechanical reasons we decided to machine the motor mounts from \SI{8}{mm} 6061-T651 aluminium. The base plate is machined from \SI{3}{mm} 6061-T6 aluminium, and the mid-level plate from \SI{2}{mm} 6061-T6. The aluminium also acts as a heat sink for controllers and motors. Even though size and arrangement of our wheels is unique, it was very helpful to use the open source designs from Tigers Mannheim~\cite{RJ20} as a reference for our base plate design.

Shells for our robots are printed using either white PolyLite ASA, or a matte black PETG. Both are more suitable as material for the shell than PLA for mechanical and thermal properties. For the top cover, we use black PLA-CF for its matte visual appearance. Shells are attached to the mid plate using M$4 \times \SI{70}{mm}$ brass standoffs and two black thumbscrews. Kicker front and battery holder suspended from the mid-level plate are printed in PLA-CF, and some of the kicker mechanics is machined from aluminium.

\begin{figure}[tbp]
\centering
\includegraphics[width=0.48\textwidth]{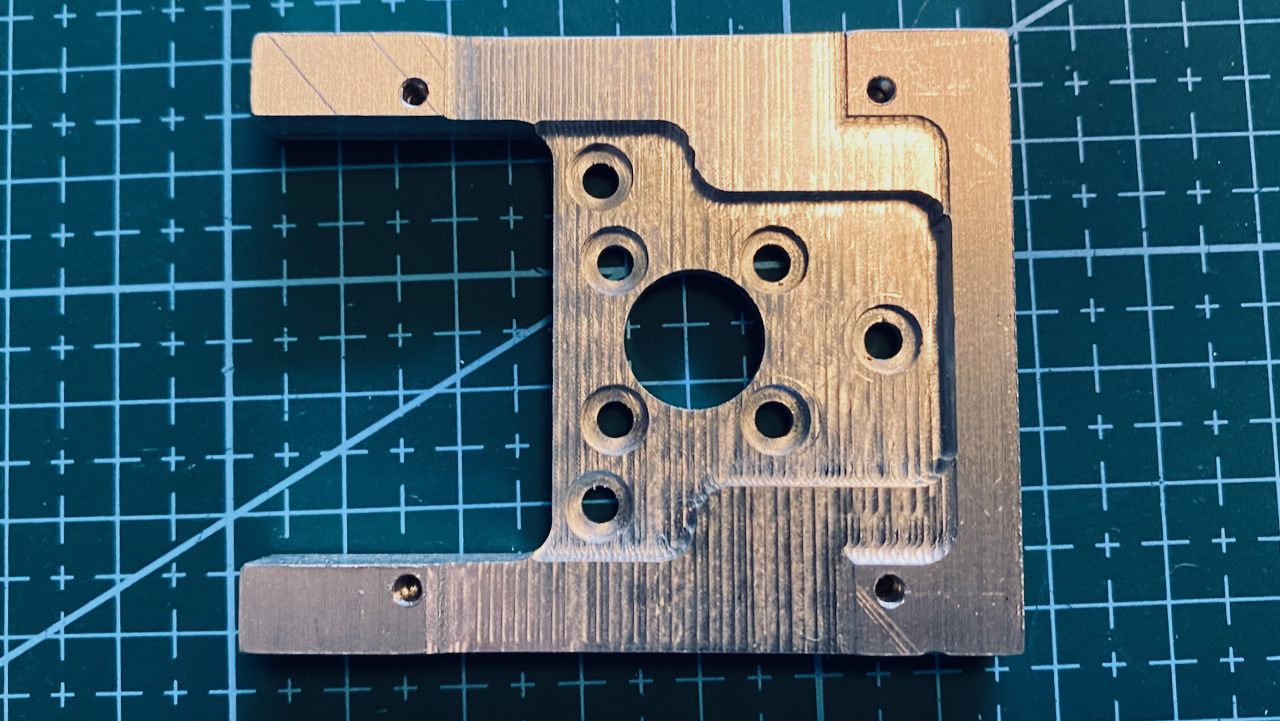}\hfill%
\includegraphics[width=0.48\textwidth]{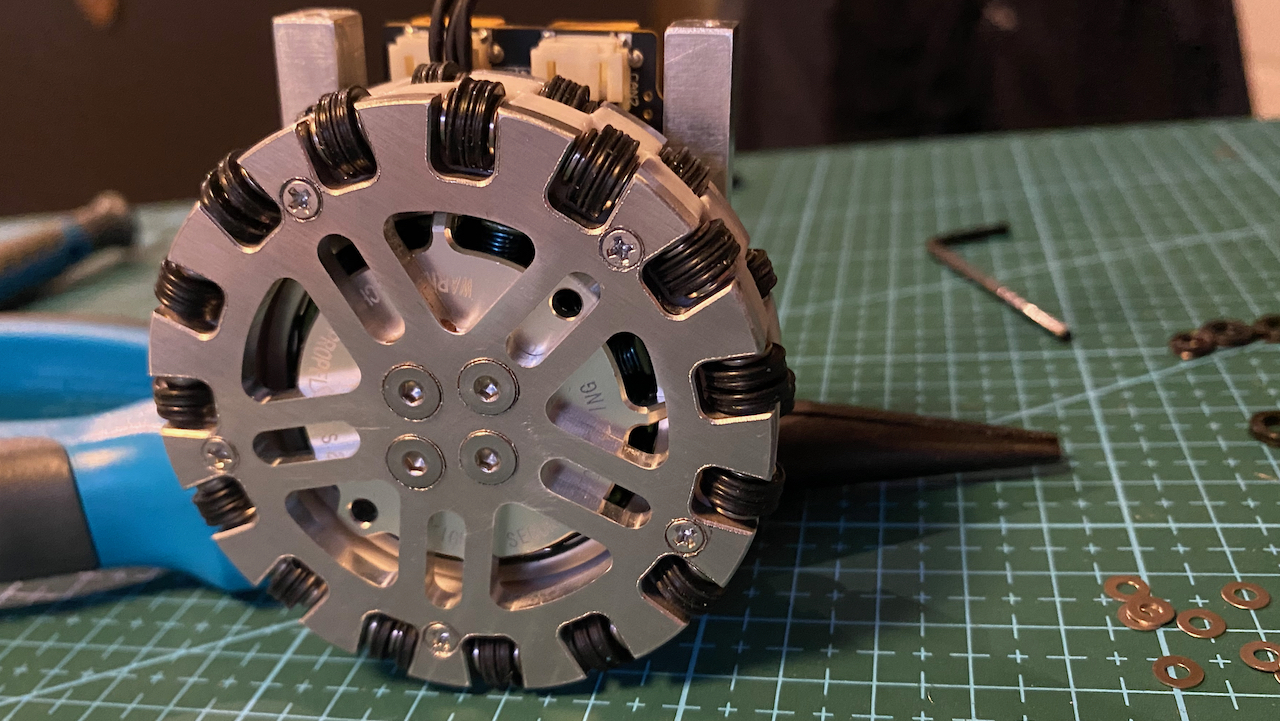}
\caption{Left: Motor mount Right: Completed drive with motor mount, wheel, and controller.}
\label{fig:motormount}
\end{figure}

Other parts required include stand-offs and various screws in M2, M2.5, M3 and M4 sizes. Our dribbler is currently still under development, and we are planning to experiment with different kinds of tubing to investigate suitable materials, similar to the KIKS team study in~\cite{MNM+23}. It is clear from a few approaches (e.g., \cite{AAL+19,HWC+19,SSB+23}) that both geometry and material selection provide a lot of space for experimentation and design, something we have planned for the time before the competition. 

\begin{table}[t]
\caption{Mechanical components used for one robot. Prices are approximate, per robot. Required aluminium sheet and bar sizes are approximate and dependent on supplier and CNC machine sizes. Total mechanical costs do not include printed parts, machine time, tools, screws, or shipping; for a single robot the listed mechanical components add up to approx.\ US\$179 (or approx.\ AU\$275).}
\centering
\begin{tabular}{rllr}
\toprule
\multicolumn{2}{l}{Component} & Specs & Price (US\$) \\ \midrule
120 & dowel pins & $\SI{2}{mm} \times$\SI{10}{mm}, stainless steel & 3.40 \\
240 & MF52 2Z & $2\times5\times2.5$ flanged bearing & 66.00 \\
240 & NBR X-rings & ID $\SI{4.47}{mm} \times \SI{1.78}{mm}$ & 47.00 \\
40 & heat set inserts & M$2 \times \SI{4}{mm} \times \SI{3.2}{mm}$ & 1.00 \\
8 &  wheel covers & ${100} \times {100} \times \SI{3.18}{mm}$ 6061-T6 aluminium & 18.00\\
4 &  motor mounts & $62 \times 50 \times \SI{7.94}{mm}$ 6061-T651 aluminium & 26.60\\
4 & wheel chassis, 3D printed & PETG & \\
\multicolumn{2}{l}{Base plate} & $200 \times 200 \times \SI{3.18}{mm}$  6061-T6 aluminium &  9.00 \\
\multicolumn{2}{l}{Mid plate} & $200 \times 200 \times \SI{2.03}{mm}$  6061-T6 aluminium &  8.00 \\
\multicolumn{2}{l}{shell, 3D printed} & ASA or PETG & \\
\multicolumn{2}{l}{shell top and kicker, 3D printed} & PLA-CF & \\
\bottomrule
\end{tabular}
\label{tab:mechanical}
\end{table}%

\section{Field setup and communication}
\label{sec:field}

The official playing area size for SSL Division B is 9 m by 6 m. For practical and cost saving reasons, we set up a smaller field size of approximately 5 m by 2.75 m. Our room has a relatively low usable ceiling, of approximately 2.65 m, and the chosen field size allowed us to work with a single camera. As the surface, we use a DIY polypropylene golf putting carpet from a local hardware store (US\$230 or A\$350). We tried multiple cameras: (1) an ELP wide-angle global shutter camera from AliExpress (114 degrees field of view), for approx US\$90 (A\$138), and (2) a StereoLabs Zed 2i camera (120 degrees field-of-view) for approximately US\$500 (A\$800). Both cameras work with v4l drivers on Ubuntu; we tested both at 60Hz, though both also advertise higher supported frame rates (90Hz and 100Hz, respectively). We decided to use one of the outputs of the stereo camera for our current setup, as the barrel distortion for the cheaper AliExpress camera is quite significant, while the outputs from the Zed camera appear almost undistorted in comparison.    

We are planning to experiment with larger field sizes, multiple cameras, and virtual camera splitters for the stereo output, in order to re-create a more faithful standard ``Division B'' setup, as well as to increase quality of the vision system output. 

At present, we use the Raspberry WLAN for communication between robots and team controller. Earlier version of the \mjb{} pi3hat featured a spread spectrum nRF24L01 interface, but unfortunately recent versions of the pi3hat no longer have this component. To provide more options for communication in radio-congested environments like RoboCup competitions, we aim to add a similar transceiver module to the Arduino instead, as one of the next steps.

\section{Software} 
\label{sec:software}

At present, the various software modules are the components undergoing most development in our team. We started development of the team mid May 2023. A significant fraction of  the time that we spent in the 9 months since then went into team organisation, hardware design, and manufacturing of our robots. Nevertheless, the software is at a stage where our robots can autonomously follow or intercept a ball, and avoid obstacles detected by the global vision system. The software is not yet at a stage where it autonomously follows the rules of the game, and there is no component that implements a strategy or coordination between individual robots yet.  We do have software modules running locally on the Raspberry Pis that implement our motor control, and a first level of team control running on a central machine using radio / WiFi communication (see also Sec.~\ref{sec:field}) with the robots. We also implemented preliminary tools to aid further development of our software.
The architecture we used to record the qualification video is based on an agent-based approach where behaviours can be replaced on demand. For all our code we used python as programming language. 

\subsection{Structure}

To test and experiment with our hardware, we implemented multiple versions of motor control on the Raspberry Pis. The purpose of our ``motor control'' is to translate requests for a robot velocity into individual motor velocities. As mentioned earlier, the moteus boards implement different control loops out of the box, including a velocity control mode. Our motor control needs to set the desired motor velocities for each of the four motors. Our program receives the requested robot velocities, calculates the appropriate motor velocities using the geometry of the robot, and passes these resulting values to the moteus controllers via CAN bus. Any lower level control is handled by the moteus controllers and firmware. Our python module also handles communication with the centralised team controller. 

As one of the first tests, we were also able to implement a ``remote-control'' mode, where robots instantly follow keyboard commands on a computer connected via WiFi.  

\subsection{Path Planner}
The path planner used in this study implements a Probabilistic Roadmap (PRM) approach~\cite{KSLO96}, adapted from the open source repository (\href{https://github.com/KaleabTessera/PRM-Path-Planning}{https://github.com/\allowbreak KaleabTessera/PRM-Path-Planning}). The algorithm consists of several steps aimed at achieving safe and efficient robot navigation in dynamic environments. First, a set of $n$ random samples, called milestones, is generated within the configuration space. These milestones serve as reference points for constructing a roadmap of feasible paths. A collision check is then performed to determine the feasibility of these milestones, ensuring that they do not intersect with obstacles or other robots present in the environment. 

Importantly, the positions of obstacles are extracted from the SSL vision system and fed into the algorithm. This integration allows the path planner to dynamically adapt to changes in the environment, ensuring accurate representation of obstacle positions and facilitating collision-free path planning. Once collision-free milestones have been identified, the algorithm proceeds to find the $k$ valid neighbours for each milestone. By connecting each milestone to its $k$ nearest neighbours, links are generated. These links are only retained as local paths if they are collision free.

In addition, the algorithm incorporates a critical parameter -- the number of samples used to generate milestones. This parameter plays a key role in balancing path efficiency and computational overhead. While a higher number of samples generally leads to more efficient paths by providing a finer representation of the configuration space, it also leads to increased computational overhead. To address this trade-off, experiments were conducted to determine an optimal value for the number of samples. In our implementation, the use of 10 samples proved to be an effective compromise, allowing the generation of diverse and collision-free paths while efficiently managing computational resources.

\begin{figure}[tbp]
\centering
\includegraphics[width=\textwidth]{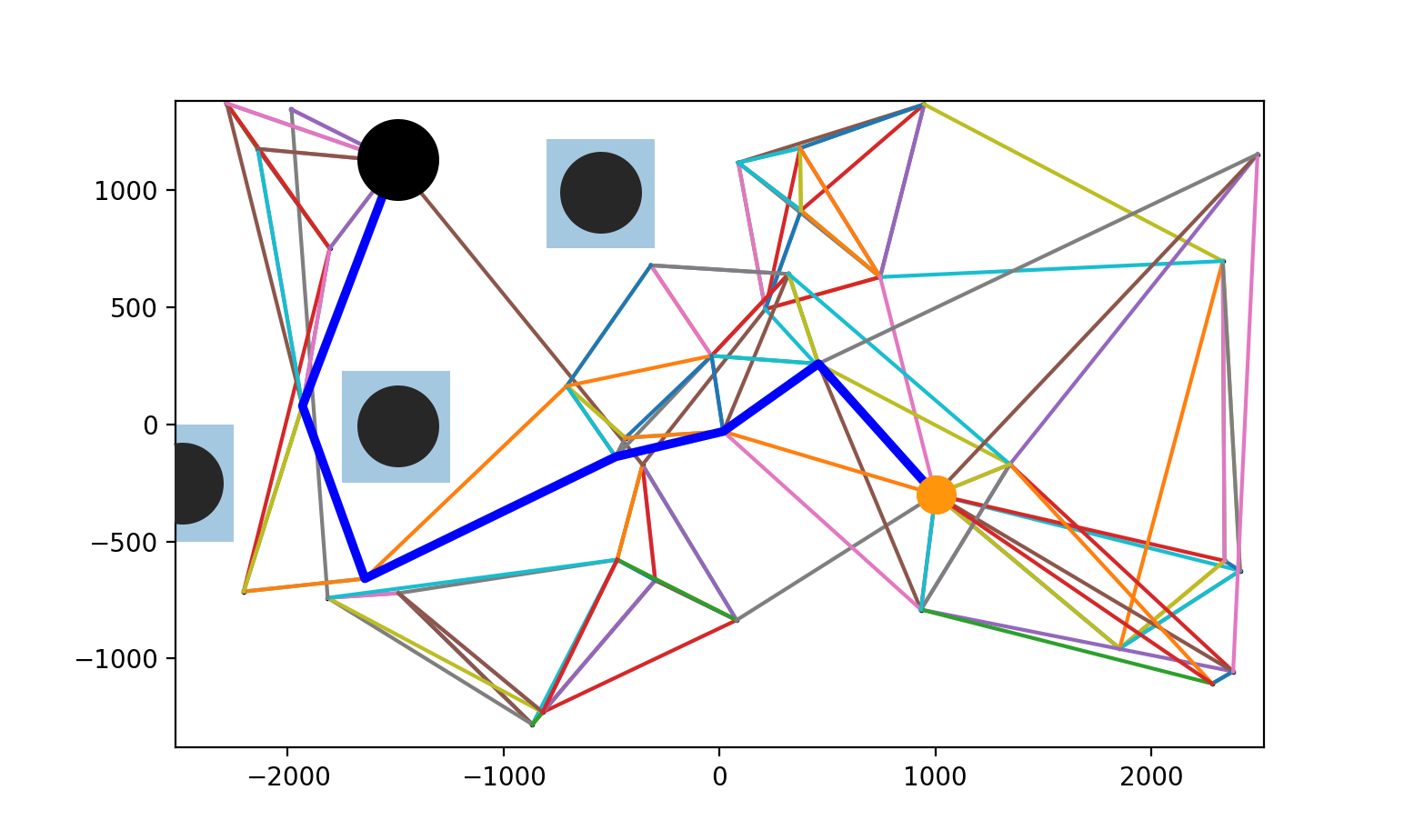}
\caption{Path planner example for a robot moving to the ball with 20 random milestones. The other robots are seen as obstacles. While the thin lines show collision free paths between neighbouring milestones, the thick blue lines show the shortest path.}
\label{fig:pathplanner}
\end{figure}

Finally, Dijkstra's shortest path algorithm~\cite{dijkstra1959note} is used to find the shortest path from the start (acting robot) to the end node (target position, e.g. the ball) within the PRM graph (see Fig.~\ref{fig:pathplanner}). Our approach is simple and easy to implement, making it well suited as an initial path planner.

We added code to call the planner only when necessary: if there is an obstacle-free direct path between current position and target, we remove any active plan and just move towards the target. If the direct path is obstructed, and if there is an active plan, we check if we can still reach the next milestone without running into an obstacle. Only if this is not the case, we will create a new plan.

\subsection{Trajectory Estimation}
\begin{figure}[tbp]
\centering
\begin{subfigure}{0.8\textwidth}
  \centering
  \includegraphics[width=\linewidth]{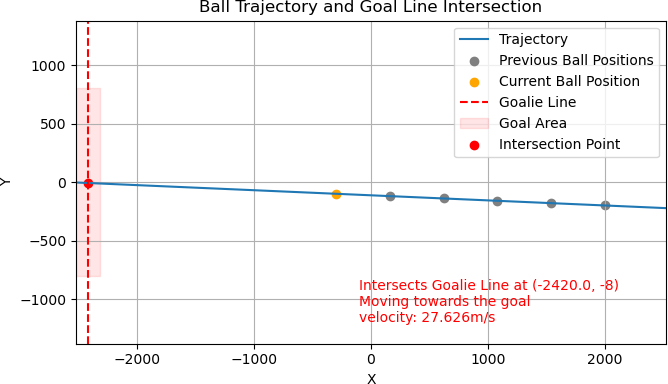}
  \caption{Example for a the ball trajectory where the ball does go into the goal.}
  \label{fig:sub1}
\end{subfigure}
\begin{subfigure}{0.8\textwidth}
  \centering
  \includegraphics[width=\linewidth]{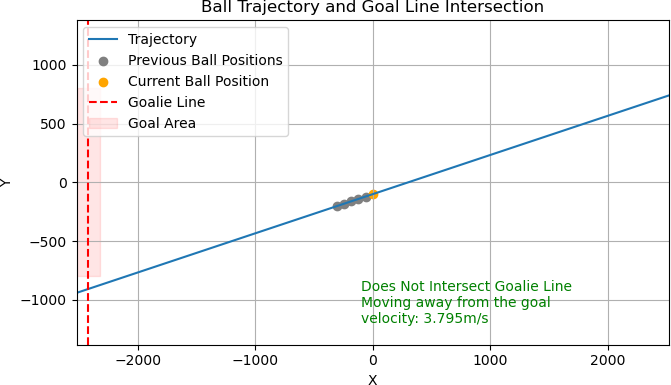}
  \caption{Example for a ball trajectory where the ball does not go into the goal.}
  \label{fig:sub2}
\end{subfigure}
\caption{Examples of ball trajectory estimation based on linear regression.}
\label{fig:balltrajectory}
\end{figure}

For the trajectory estimation of the ball we use a linear regression model.
We use linear regression to predict the trajectory of a moving ball. The system uses a series of observed ball positions obtained from the SSL vision software. By fitting a linear regression model to the last few frames of ball positions, our system estimates the future trajectory of the ball. It then assesses whether this trajectory intersects with the line along which the goalkeeper is moving, indicating that the opponent will score. In addition, the framework provides essential information such as the direction and speed of the ball's movement, the intersection of the ball's trajectory with the goalkeeper's trajectory (see Fig.~\ref{fig:balltrajectory}), so that the goalkeeper can move into position to block the ball. Furthermore, the estimated trajectory can be used by other robots trying to receive a pass or steal the ball. 

The main reason for using linear regression for trajectory estimation in our software is its simplicity and ease of implementation.

\subsection{Strategy}
In the future, our game will feature a dynamic system where each robot is assigned a specific role from a pre-defined list at the start of the match. 
So far, we implemented a strategic positioning module that 
keeps track of different formations and roles the different robots will take. It defines, for each formation, a dynamically calculated ``home'' position for each robot, dependent on ball position and player role. 
The roles influence how much the robot position will change with the position of the ball, and also if the robot will, in general, aim to stay behind the ball.
There is also a formation manager module that can read the different formations from a plain text file. 

We ran simulations with our strategic positioning module, but have not tested on robots yet. It will add an element of strategy to the gameplay, and will also be useful for setplays. The available roles include the goalkeeper, though this role is possibly less influenced by the formation than other players. The goalie will be responsible for guarding the goal area by moving along a vertical line directly in front of the goal. With our formations and roles module we can realise  different types of defenders, who will be responsible for intercepting passes and shots towards the goal from the opposing team; a main attacker, who will focus on scoring goals and passing as needed; and the attacker's assistants, who will position themselves to receive passes from the main attacker and potentially transition into the primary scoring role themselves. This role allocation mechanism promises to enhance the tactical dimension of the game, providing players with varied responsibilities and opportunities for strategic coordination.

\section{Conclusion}
\label{sec:conclusion}

This paper describes our first attempt in qualifying for RoboCup SSL league. One aim of this report is to give an approximate overview of hardware costs to establish a new SSL team, with what we believe is a solid but cost-efficient approach that leverages more off-the-shelf hardware than most other teams. Teams in Division B have up to 6 robots on the field. With an approximate cost of US\$1,230 (A\$1,900) for each of our robots, and US\$730 / A\$1,150 for the field setup (excluding a PC to run the vision software), total initial cost for a team is about  US\$8,110 (A\$12,500). 
This amount does not include substitute robots, necessary reserves for replacements of components that inevitably will get damaged as part of the process, nor tools and other minor expenses. For our qualification, we have completed two robots (with the dribbler still work in progress), and an earlier robot prototype with a plywood base and 3D-printed motor mounts.

With our open-source hardware design and an approach that bypasses some of the complexity inherent in the designs of more established teams, we aspire to lower the barriers to entry for the league. This is particularly aimed at those starting in the league with a smaller emphasis on electronics engineering. Our strategy emphasises accessibility and educational value. Our github repositories for our hard- and software are at \href{https://github.com/WSU-TurtleRabbit}{https://github.com/WSU-TurtleRabbit}.

\section*{Acknowledgements}

We extend our gratitude to Pradeshi Khadka for contributions during the early stages of our project. Additionally, we acknowledge the support provided by the School of Computer, Data, and Mathematical Sciences for parts of our hardware. Our thanks also go to the Western Sydney University Robotics, Automation and Manufacturing (RAM) Club for their administrative support.

%
%
%
\bibliographystyle{splncs04}
\bibliography{turtlerabbit}
\end{document}